\documentclass[letterpaper, 10 pt, conference]{ieeeconf}  

\IEEEoverridecommandlockouts                              

\usepackage{amsmath} 
\usepackage{amssymb}  
\usepackage{graphicx}
\usepackage{caption}
\usepackage{booktabs}
\usepackage{float}
\usepackage{cite}
\usepackage{color}
\usepackage[table]{xcolor}
\usepackage{multirow}
\usepackage{fontawesome5}
\usepackage{algorithm}
\usepackage{algorithmic}
\usepackage{hyperref}
\usepackage{fancyhdr}
\usepackage{xspace}
\usepackage{pifont}

\definecolor{darkgreen}{RGB}{34,139,34}

\newcommand{\our}{CAR\xspace}

\title{\LARGE \bf
\our: Cross-Vehicle Kinodynamics Adaptation via \\ Mobility Representation
}

\author{Tong Xu$^{*}$, Chenhui Pan$^{*}$, and Xuesu Xiao
\thanks{All authors are with the Department of Computer Science, George Mason University {\tt\small \{txu25, cpan7, xiao\}@gmu.edu}}
\thanks{*Equally contributing authors}
}

\begin{document}

\maketitle
\thispagestyle{empty}
\pagestyle{empty}

\begin{abstract}
Developing autonomous mobile robot systems typically requires either extensive, platform-specific data collection or relies on simplified abstractions, such as unicycle or bicycle models, that fail to capture the complex kinodynamics of diverse platforms, ranging from wheeled to tracked vehicles. This limitation hinders scalability across evolving heterogeneous autonomous robot fleets. To address this challenge, we propose Cross-vehicle kinodynamics Adaptation via mobility Representation (\our), a novel framework that enables rapid mobility transfer to new vehicles. \our employs a Transformer encoder with Adaptive Layer Normalization to embed vehicle trajectory transitions and physical configurations into a shared mobility latent space. By identifying and extracting commonality from nearest neighbors within this latent space, our approach enables rapid kinodynamics adaptation to novel platforms with minimal data collection and computational overhead. We evaluate \our using the Verti-Bench simulator, built on the Chrono multi-physics engine, and validate its performance on four distinct physical configurations of the Verti-4-Wheeler platform. With only one minute of new trajectory data, \our achieves up to 67.2\% reduction in prediction error compared to direct neighbor transfer across diverse unseen vehicle configurations, demonstrating the effectiveness of cross-vehicle mobility knowledge transfer in both simulated and real-world environments.

\end{abstract}
\section{Introduction}
Kinodynamics modeling across heterogeneous vehicle platforms remains a fundamental challenge in mobile robotics~\cite{xiao2022motion, wang2024survey}. Modern ground robot fleets encompass vehicles with diverse physical configurations, including variations in suspension stiffness, drivetrain architecture, payload distribution, and actuation type, ranging from wheeled to tracked systems. These structural differences lead to distinct kinodynamics behaviors, even under similar operating conditions. To address such diversity, prior works have proposed platform-specific mobility solutions through physics-based modeling~\cite{cai2025pietra, zhao2024physord}, learning-based kinodynamics models~\cite{xiao2021learning, datar2024terrain, atreya2022high, pokhrel2024cahsor}, and end-to-end policies~\cite{xu2025vertiselector, pan2020imitation}. While these approaches have demonstrated promising results on individual platforms, they are typically designed and validated for a single vehicle type. As a result, they lack a holistic framework that enables different ground vehicles to share and transfer common kinodynamics knowledge across a heterogeneous robot fleet.

In practice, when a new vehicle is introduced or an existing platform is modified, conventional approaches require collecting large-scale, platform-specific datasets and retraining mobility models from scratch~\cite{borges2022survey}. This repetitive engineering process is both time-intensive and computationally expensive, fundamentally constraining the scalability of fleet deployments. Therefore, a scalable framework is required to enable knowledge sharing across the fleet while preserving the flexibility to rapidly adapt to platform-specific kinodynamics.

\begin{figure}[t!]
    \centering
    \includegraphics[width=\columnwidth]{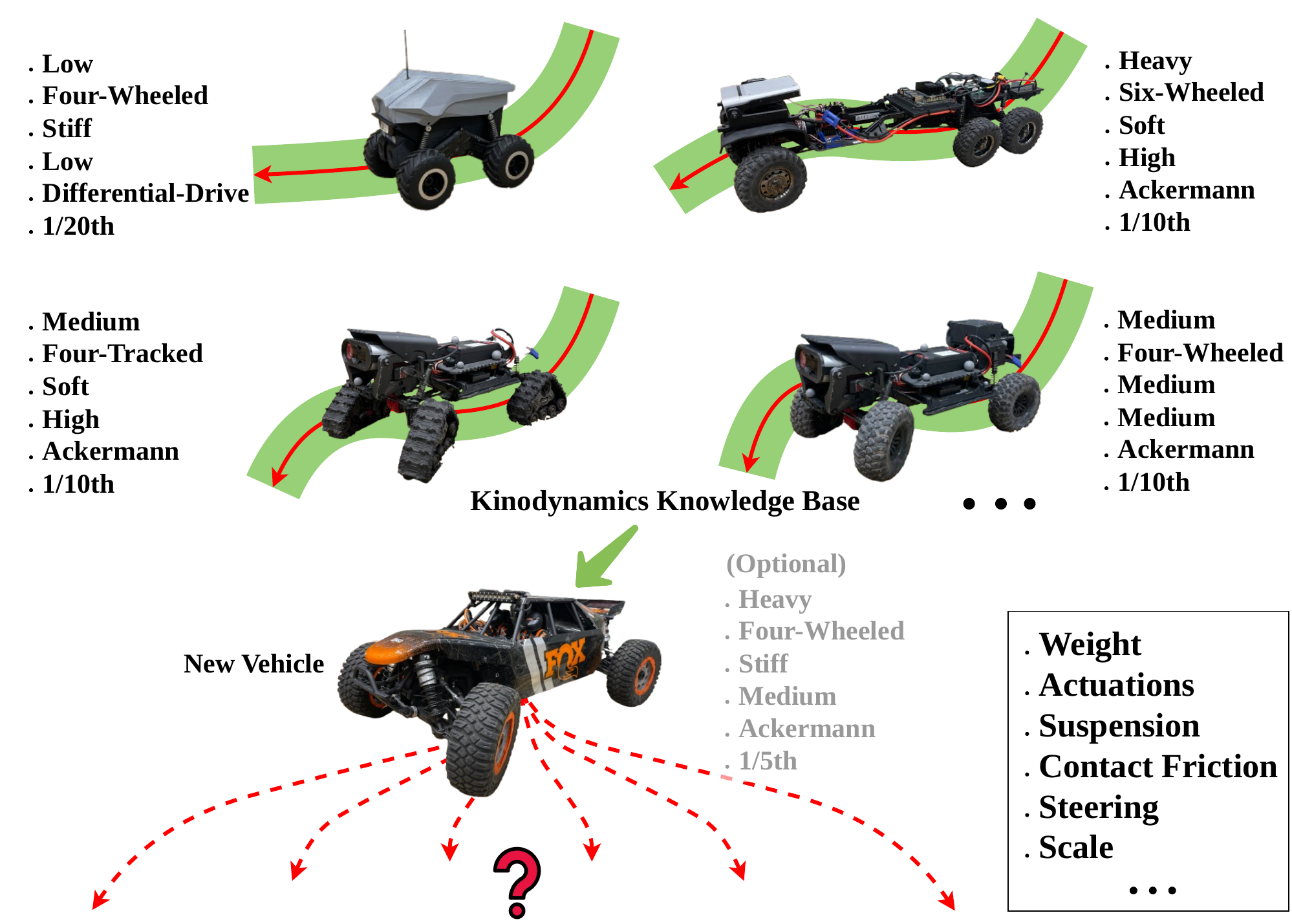}
    \caption{A heterogeneous fleet of vehicles with diverse physical configurations form a kinodynamics knowledge base. When a new vehicle is introduced, shared mobility representations in this knowledge base enable rapid adaptation with minimal data, without designing or retraining from scratch.} 
    \label{fig::motivation}
\end{figure}

To overcome this scalability limitation and enable adaptation across heterogeneous vehicle platforms, self-supervised representation learning offers a promising direction by uncovering shared structure in mobility data without explicit supervision. In particular, contrastive learning methods have demonstrated strong capability in constructing structured latent space, where semantically similar samples are encouraged to cluster while dissimilar ones are separated~\cite{chen2020simple, he2020momentum}. Such representations provide transferable priors that generalize across diverse tasks and domains. In robotics, learned representations have been leveraged for task transfer~\cite{finn2017model, devin2017learning}, sim-to-real adaptation~\cite{tobin2017domain, peng2018sim}, and multi-task or multi-skill policy learning~\cite{haarnoja2018latent, yu2020meta}. However, learning cross-vehicle kinodynamics representations that effectively incorporate various physical configurations to adapt to new vehicles with minimal data remains an open challenge.

Motivated by such a challenge, we present \textbf{C}ross-vehicle kinodynamics \textbf{A}daptation via mobility \textbf{R}epresentation (\our), a novel framework that enables rapid kinodynamics adaptation to new vehicle platforms through a shared mobility knowledge base across a heterogeneous fleet, as shown in Fig.~\ref{fig::motivation}. Our contributions are summarized as follows:

\begin{itemize}
    \item A cross-vehicle mobility representation framework that employs a Transformer encoder with Adaptive Layer Normalization (AdaLN)~\cite{peebles2023scalable} to embed trajectory transitions and physical configurations into a structured, shared latent space; 
    \item An efficient selection strategy that identifies relevant mobility neighbors within the shared latent space to extract transferable kinodynamics knowledge in order to minimize data overhead on new vehicles;
    \item A rapid adaptation mechanism that integrates weighted dataset aggregation, weighted loss optimization, and regulated gradient update to transfer mobility knowledge to new vehicles using only one minute of new trajectory data; and
    \item Empirical validation of our method in simulation~\cite{xu2025verti} and on four distinct physical configurations of the Verti-4-Wheeler platform~\cite{datar2024toward}, demonstrating superior performance over state-of-the-art baselines.
\end{itemize}

\section{Related Work}
In this section, we review related work on kinodynamics modeling, self-supervised representation learning, and cross-vehicle knowledge transfer for rapid adaptation.

\subsection{Kinodynamics Modeling}
Kinodynamics modeling seeks to capture the coupled relationship among various vehicular components to enable accurate motion prediction and planning~\cite{cai2025q}. Traditional approaches relied on physics-based models that explicitly parameterize vehicle properties such as size, weight, friction, actuation, suspension, and contact forces~\cite{pavlov2019soil}. While physically interpretable, these models require careful parameter tuning and often struggle to generalize across different platforms.

More recently, learning-based approaches have emerged as a compelling alternative, leveraging data-driven methods to approximate kinodynamics behavior from onboard sensory observations~\cite{meng2023terrainnet, pokhrel2024cahsor}. End-to-end policy learning methods further bypass explicit kinodynamics modeling by directly learning control policies from demonstration data~\cite{xu2024reinforcement}. Despite these advances, existing approaches remain platform-specific and do not generalize across varying vehicle configurations. The challenge of learning shared mobility representations that transfer across heterogeneous platforms remains unaddressed.

\subsection{Representation Learning in Robotics}
Self-supervised representation learning has emerged as a powerful paradigm for extracting structured and transferable representations from high-dimensional robotic data. Contrastive learning methods~\cite{chen2020simple, he2020momentum} construct structured latent space by aligning representations of positive augmented views while separating dissimilar samples. Beyond contrastive approaches, cross-embodiment learning has explored the transfer of representations and policies across robots with diverse physical configurations~\cite{Yang2024Pushing, doshi2025scaling}. These approaches typically learn shared representations that abstract embodiment-specific details while preserving task-relevant structure, enabling policy reuse across platforms with varying kinematics or actuation layouts. However, existing cross-embodiment approaches primarily focus on policy-level behavior learning despite embodiment differences, rather than explicitly modeling distinct kinodynamics grounded in diverse physical configurations. In other words, they \emph{abstract away} embodiment differences, rather than \emph{adapt to} them. 

\subsection{Rapid Adaptation and Knowledge Transfer}
Few-shot adaptation aims to rapidly generalize a learned model to new tasks or domains using minimal data. Foundational approaches, such as Model-Agnostic Meta-Learning (MAML)~\cite{finn2017model} and its first-order extensions~\cite{nichol2018first, fallah2020convergence, nagabandi2019learning}, achieve this by optimizing parameter initializations that can be efficiently fine-tuned with only a few gradient updates. In robotics, AnyCar~\cite{xiao2025anycar} tackles a similar adaptation problem by learning a universal dynamics model that generalizes across multiple vehicle platforms. However, they are usually incapable of knowing when adaptation is not possible due to immitigable differences of a new task. Furthermore, despite their applicability to a wide range of tasks, these purely data-driven methods do not leverage problem-specific knowledge, e.g., vehicle configurations, to efficiently facilitate adaptation. In extremely data-scarce scenarios where even few-shot fine-tuning is not possible, such adaptation efficiency is paramount. One avenue toward adaptation efficiency is to only identify and leverage prior knowledge closely relevant to a new task, while effectively mitigating its difference with existing ones.  

Therefore, \our moves beyond the limitations of few-shot adaptation by explicitly leveraging kinodynamics-specific vehicle configurations and trajectories to identify the most relevant mobility neighbors within a structured latent space, excluding irrelevant vehicles to prevent unnecessary and detrimental adaptation. By combining weighted dataset aggregation and loss optimization with gradient updates regulated by extremely scarce new vehicle data, our framework enables rapid kinodynamics adaptation to new vehicle configurations using only one minute of new trajectory data.

\section{Method}

\begin{figure*}[ht]
    \centering
    \includegraphics[width=2\columnwidth]{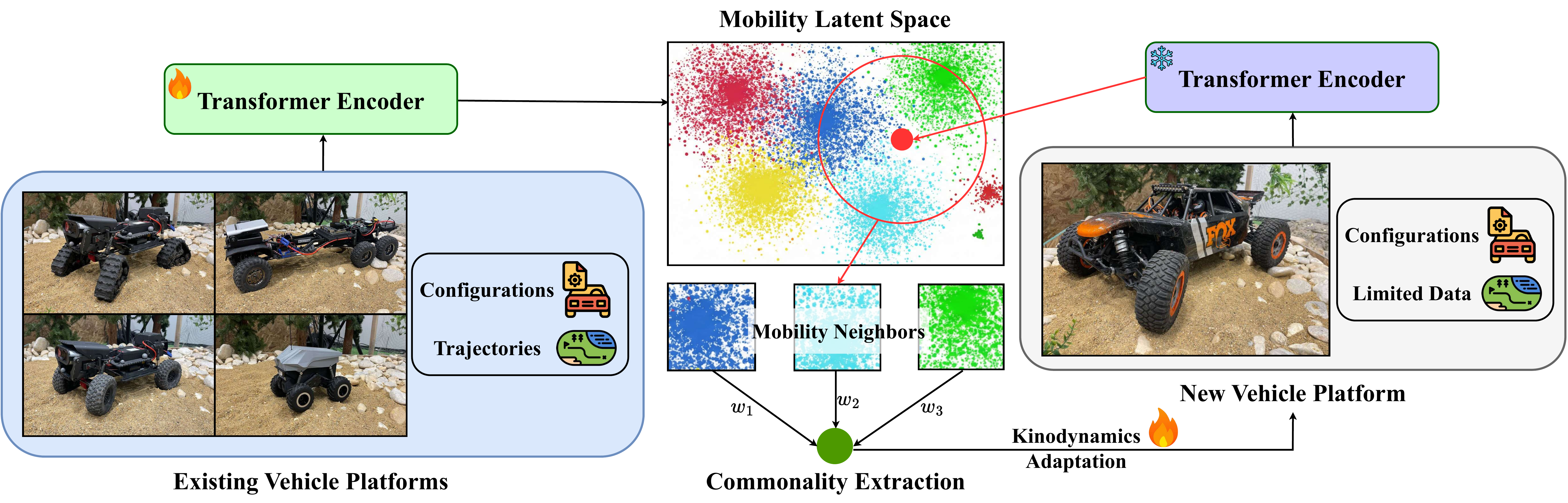}
    \caption{\our Overview: Physical configurations and vehicle trajectories from a diverse fleet (left) are embedded into a shared mobility latent space (middle). When a new vehicle is introduced (right), its representation identifies nearest mobility neighbors in this mobility latent space. Inversely proportional to the distances to the mobility neighbors, weights $w_1, w_2, w_3$ are used to aggregate datasets and bias training loss. Gradient update regulated by minimal new vehicle data then enables rapid kinodynamics adaptation.}
    \label{CAR}
\end{figure*}

Our objective is to enable rapid kinodynamics adaptation to new vehicle platforms with minimal new data. We decompose our method into three phases: (1) learning a shared mobility latent space that encodes physical configurations and vehicle trajectories, (2) identifying the most relevant mobility neighbors within this space, and (3) leveraging prior knowledge to enable rapid kinodynamics adaptation. An overview of the framework is illustrated in Fig.~\ref{CAR}.

\subsection{Problem Formulation}
\label{sec::pf}
We define vehicle kinodynamics as a six Degrees of Freedom (DoFs) problem over the state space $S \subset \mathbb{SE}(3)$. The vehicle state at time $t$ is defined as:
\begin{equation*}
    \mathbf{s}_t = \big[x_t, y_t, z_t, \phi_t, \varphi_t, \eta_t\big] \in \mathbb{SE}(3),
\end{equation*}
where $x_t, y_t, z_t$ represent the vehicle position in 3D, and $\phi_t, \varphi_t, \eta_t$ denote roll, pitch, and yaw. Higher-order derivatives can be included in $\mathbf{s}_t$ to capture more dynamic maneuvers (see experiments for details). The control input $\mathbf{u}_t \in \mathcal{U}$ corresponds to steering and speed. The forward kinodynamics model of a single vehicle is expressed as:
\begin{equation*}
    \mathbf{s}_{t+1} = f(\mathbf{s}_t, \mathbf{u}_t).
\end{equation*}

Traditional kinodynamics models are typically trained independently for each vehicle platform and do not explicitly account for variations in physical configurations, limiting their transferability within a heterogeneous fleet. We represent the physical configuration of a vehicle $v$ as a parameter descriptor $\mathbf{c} \in \mathbb{R}^{d_c}$, where $d_c$ denotes the number of configurable physical parameters. This descriptor encodes platform-specific properties that govern kinodynamics behavior, such as chassis mass, suspension stiffness, tire friction, actuation layout, and geometric scale.

To enable cross-vehicle knowledge sharing, we assume access to an existing heterogeneous fleet $\mathcal{V}_{\text{train}} = \{v_1, v_2, \dots, v_k\}$. For each vehicle $v_i \in \mathcal{V}_{\text{train}}$, characterized by a distinct physical configuration $\mathbf{c}_i$, a trajectory dataset $\mathcal{D}_i = \{\tau_1, \tau_2, \dots, \tau_M\}$ is available as prior knowledge, e.g., from previous design effort or deployment experiences. Each trajectory $\tau_j = \{(\mathbf{s}_t, \mathbf{u}_t, \mathbf{s}_{t+1})\}_{t=1}^{H}$ consists of $H$ discrete kinodynamics transition steps. 

Given a new vehicle $v_{\text{new}}$, the problem becomes to derive the forward kinodynamics model $f_{\theta}$, parameterized by $\theta$,  with as little new data as possible, i.e., $\mathcal{D}_\textrm{new} = \{\tau_1, \tau_2, \dots, \tau_{M_\textrm{new}\}}, M_\textrm{new} < < M$, with or without its physical configuration $\mathbf{c}_\textrm{new}$.

\subsection{Cross-Vehicle Mobility Representation}
\label{sec::cross_latent}
Given trajectory transitions and physical configurations from a heterogeneous training fleet, we aim to learn an encoder that maps each vehicle's mobility behavior into a shared latent space. This latent space should (i) preserve kinodynamics (dis)similarity across vehicles and (ii) incorporate physically meaningful priors from vehicle configurations to support downstream mobility neighbor identification and rapid kinodynamics adaptation.

\subsubsection{Physical Configuration Embeddings}
\label{sec:param}
The physical parameters comprising $\mathbf{c}$, such as chassis mass, tire friction, and suspension stiffness, are inherently continuous quantities that characterize the kinodynamic behavior of a vehicle. To incorporate these platform-specific physical priors, we encode $\mathbf{c}$ into a compact latent representation via a learnable encoder $g_\xi(\cdot)$:
\begin{equation*}
\mathbf{e}_c = g_\xi(\mathbf{c}) \in \mathbb{R}^{d},
\end{equation*}
where $d$ is the embedding dimension. The resulting embedding $\mathbf{e}_c$ serves as a conditioning vector that encodes physical priors for downstream kinodynamics modeling.

\subsubsection{Transformer Encoder with AdaLN Conditioning}
\label{sec:adaln}
We employ a Transformer encoder $e_{\psi}$ to embed trajectories into a shared mobility latent space. Each transition token $\mathbf{x}_t = [\mathbf{s}_t, \mathbf{u}_t, \mathbf{s}_{t+1}]$ is first projected to dimension $d$, prepended with a learnable CLS token $\mathbf{h}_{\mathrm{cls}}$, and augmented with sinusoidal positional embeddings $\mathrm{PE}$:
\begin{equation*}
\mathbf{H}^{(0)} = [\mathbf{h}_{\mathrm{cls}},\, \mathrm{Proj}(\mathbf{x}_{t_0}), \dots, \mathrm{Proj}(\mathbf{x}_{t_0+L-1})] + \mathrm{PE}.
\end{equation*}
The sequence is processed through $N$ Transformer blocks. To inject physical configuration priors without overwhelming trajectory encoding, AdaLN~\cite{peebles2023scalable} conditioning is applied only at the final block. The configuration embedding $\mathbf{e}_c$ produces per-channel affine modulation parameters via a linear projection $\Phi$:
\begin{equation*}
[\Delta\boldsymbol{\beta}_1, \Delta\boldsymbol{\gamma}_1, \Delta\boldsymbol{\beta}_2, \Delta\boldsymbol{\gamma}_2] = \Phi(\mathbf{e}_c),
\end{equation*}
which modulates the attention and feed-forward sublayers as:
\begin{align*}
\widetilde{\mathbf{H}} &= \mathbf{H}\odot (1+\Delta \boldsymbol{\gamma}_1) + \Delta \boldsymbol{\beta}_1,\\
\mathbf{H} &\leftarrow \mathrm{LN}\big(\mathbf{H} + \mathrm{Attn}(\widetilde{\mathbf{H}})\big),\\
\widetilde{\mathbf{H}} &= \mathbf{H}\odot (1+\Delta \boldsymbol{\gamma}_2) + \Delta \boldsymbol{\beta}_2,\\
\mathbf{H} &\leftarrow \mathrm{LN}\big(\mathbf{H} + \mathrm{FFN}(\widetilde{\mathbf{H}})\big),
\end{align*}
where $\odot$ denotes element-wise multiplication, $\mathrm{LN}(\cdot)$ is Layer Normalization, $\mathrm{Attn}(\cdot)$ is multi-head self-attention, and $\mathrm{FFN}(\cdot)$ is a position-wise feed-forward network.
The final mobility embedding is extracted from the CLS output:
\begin{equation*}
\mathbf{z} = e_{\psi}(\tau, \mathbf{c}) = \mathbf{H}^{(N)}_{\mathrm{cls}} \in \mathbb{R}^{d}.
\end{equation*}

\subsubsection{Dual-Path Triplet Training}
\label{sec:dualpath}
To structure the latent space such that kinodynamically similar vehicles cluster together while dissimilar ones are separated, we train the encoder with a dual-path triplet loss. For each anchor trajectory, we sample a positive trajectory from the same vehicle configuration and a negative trajectory from a different one, forming triplets $(\tau_a,\tau_p,\tau_n)$. The training objective consists of two complementary losses:

\textbf{(i) Unconditional Mobility Loss.} We encode trajectories without configuration conditioning to emphasize trajectory-level kinodynamics:
\begin{equation*}
\mathbf{z}^{u} = e_{\psi}(\tau, \varnothing).
\end{equation*}
We apply a triplet margin loss:
\begin{equation}
\mathcal{L}_{u} =
\max\Big(\|\mathbf{z}^{u}_{a}-\mathbf{z}^{u}_{p}\|_2
-\|\mathbf{z}^{u}_{a}-\mathbf{z}^{u}_{n}\|_2 + \delta,\; 0 \Big),
\nonumber
\end{equation}
where $\delta$ is a predefined threshold. 

\textbf{(ii) Conditional Mobility Loss.} We additionally encode trajectories with configuration conditioning to encourage physically grounded clustering:
\begin{equation*}
\mathbf{z}^{c} = e_{\psi}(\tau, \mathbf{c}),
\end{equation*}
and apply the same triplet loss:
\begin{equation}
\mathcal{L}_{c} =
\max\Big(\|\mathbf{z}^{c}_{a}-\mathbf{z}^{c}_{p}\|_2
-\|\mathbf{z}^{c}_{a}-\mathbf{z}^{c}_{n}\|_2 + \delta,\; 0 \Big).
\nonumber
\end{equation}
The final training objective is $\mathcal{L} = \mathcal{L}_{u} + \mathcal{L}_{c}$.
This dual-path objective jointly captures intrinsic mobility patterns of the trajectories and the influence of physical vehicle configurations, producing a structured latent space that supports reliable cross-vehicle knowledge transfer.

\subsection{Mobility Neighbor Identification}
\label{sec::mobility_neigh}
Given the trained Transformer encoder $e_{\psi}$, we project the new vehicle $v_{\text{new}}$ into the shared mobility latent space and identify its relevant mobility neighbors from $\mathcal{V}_{\text{train}}$.

\textbf{Centroid Computation.} For each vehicle $v_i \in \mathcal{V}_{\text{train}}$, we encode a set of trajectories using the conditional encoder and compute a centroid $\boldsymbol{\mu}_i$ in the latent space:
\begin{equation*}
    \boldsymbol{\mu}_i = \frac{1}{|\mathcal{Z}_i|}\sum_{\mathbf{z} \in \mathcal{Z}_i} \mathbf{z}, \quad \mathcal{Z}_i = \{e_{\psi}(\tau_j, \mathbf{c}_i) \mid \tau_j \in \mathcal{D}_i\}.
\end{equation*}

To reduce noise and preserve the most informative latent dimensions, we apply Principal Component Analysis (PCA) and retain the minimum number of principal components that explain at least 90\% of the total variance. Similarly, the new vehicle centroid $\boldsymbol{\mu}_{\text{new}}$ is computed from its limited dataset $\mathcal{D}_{\text{new}}$ using the same PCA projection.

\textbf{Mobility Neighbor Identification.} 
To characterize the inter-cluster structure of the training fleet, we compute the pairwise cosine distances between all training centroids:
\begin{equation*}
    d_{ij}=d_{\cos}(\boldsymbol{\mu}_i, \boldsymbol{\mu}_j), \forall i < j.
\end{equation*}
We then define an adaptive distance threshold:
\begin{equation*}
    \epsilon = \bar{d} + 2\sigma_d,
\end{equation*}
where $\bar{d}$ and $\sigma_d$ denote the mean and standard deviation of the inter-cluster cosine distances, respectively.

Given the centroid $\boldsymbol{\mu}_{\text{new}}$, we evaluate its cosine distance to each training centroid and identify mobility neighbors as:
\begin{equation*}
    \mathcal{N} = \left\{ v_i \in \mathcal{V}_{\text{train}} 
    \mid d_{\cos}(\boldsymbol{\mu}_{\text{new}}, \boldsymbol{\mu}_i) \le \epsilon \right\}.
\end{equation*}
If no training vehicle satisfies this criterion, the new vehicle is classified as out-of-distribution to prevent detrimental adaptation.

\textbf{Inverse-Distance Weighting.} 
For each identified mobility neighbor $v_i \in \mathcal{N}$, we assign a weight inversely proportional to its squared cosine distance to the new vehicle centroid:
\begin{equation*}
    \tilde{w}_i = \frac{1}{d_{\cos}(\boldsymbol{\mu}_{\text{new}}, \boldsymbol{\mu}_i)^2}, \quad w_i = \frac{\tilde{w}_i}{\sum_{j=1}^{|\mathcal{N}|} \tilde{w}_j}.
\end{equation*}
To emphasize the most relevant neighbors, we sort neighbors in descending order of their weights and select the top neighbors whose cumulative weight exceeds $0.9$:
\begin{equation*}
    \mathcal{N}_{\text{new}} 
    = \left\{ v_{i} \mid 
    \sum_{j=1}^{i} w_{j} \ge 0.9 \quad \& \quad \sum_{j=1}^{i-1} w_{j} < 0.9 \right\},
\end{equation*}
The weights corresponding to $\mathcal{N}_{\text{new}}$ are then used to guide dataset aggregation and loss optimization during rapid kinodynamics adaptation.

\subsection{Rapid Kinodynamics Adaptation}
\label{sec::rapid_adapt}

Given the identified mobility neighbors $\mathcal{N}_{\text{new}}$ and their associated weights $w_i$, \our computes the kinodynamics model $f_{\theta}$ of the new vehicle through weighted knowledge transfer with gradient regulation.

\textbf{Weighted Dataset Aggregation.} 
For each neighbor vehicle $v_i \in \mathcal{N}_{\text{new}}$, we sample trajectories from its dataset $\mathcal{D}_i$, with the number of samples proportional to its weight $w_i$. This ensures that vehicles with higher mobility similarity contribute more during adaptation. The resulting aggregated training set is
\begin{equation*}
    \mathcal{D}_{\text{agg}} 
    = \bigcup_{v_i \in \mathcal{N}_{\text{new}}} 
    \mathcal{D}_i^{(w_i)},
\end{equation*}
where $\mathcal{D}_i^{(w_i)}$ denotes the weighted subset sampled from $\mathcal{D}_i$.

\textbf{Weighted Loss Optimization.} 
The kinodynamics prediction loss is computed separately on each neighbor dataset and combined as a weighted objective:
\begin{equation*}
    \mathcal{L}_{\text{train}} 
    = \sum_{v_i \in \mathcal{N}_{\text{new}}} 
    w_i \, \mathcal{L}(f_{\theta}; \mathcal{D}_i^{(w_i)}),
\end{equation*}
where $\mathcal{L}$ denotes the Mean Squared Error (MSE) over sampled trajectories. This objective prioritizes model updates from neighbors with the most mobility relevance to the new vehicle.

\textbf{Gradient Regulation via Limited New Data.} 
To ensure that adaptation remains consistent with the limited trajectory data $\mathcal{D}_{\text{new}}$, we regulate gradient updates using $\mathcal{D}_{\text{new}}$ as a constraint. Inspired by Gradient Episodic Memory~\cite{liu2021lifelong}, we compute constraint gradients from $\mathcal{D}_{\text{new}}$ and enforce:
\begin{equation*} 
    \nabla_{\theta}\mathcal{L}(f_\theta; \mathcal{D}_\textrm{new}) \cdot \nabla_{\theta}\mathcal{L}_{\text{train}} \ge 0.
\end{equation*}
If violations occur, the training gradient is projected onto the nearest gradient direction that does not increase the loss on $\mathcal{D}_{\text{new}}$. The constraint gradients are periodically refreshed as the model evolves.

Collectively, these three components enable \our to transfer shared knowledge from the most relevant mobility neighbors to the new platform using a small amount of new data.

\section{Implementations}
In this section, we present implementation details of our
approach and experiments.

\subsection{Vehicle Configurations and Datasets}
We evaluate \our using the Verti-Bench simulator~\cite{xu2025verti}, where trajectories are generated for a heterogeneous fleet of vehicles. In our experiments, the physical configuration of each vehicle $v$ is represented as: 
$
    \mathbf{c} = [\alpha_m, \mu_f, \alpha_s]^\top,
$
where $\alpha_m$ denotes the chassis mass scaling ratio relative to the nominal mass in the simulator, $\mu_f$ is the rigid tire friction coefficient, and $\alpha_s$ represents the suspension spring stiffness scaling ratio relative to the nominal stiffness. The training fleet comprises eight configurations sampled from continuous parameter ranges: $\alpha_m \in [0.5, 4.0]$, $\mu_f \in [0.6, 0.9]$, and $\alpha_s \in [0.6, 1.8]$.

To collect training trajectories, each vehicle executes sinusoidal random exploration on flat terrain. Steering follows $\mathbf{u}_{\text{steer}}(t)=\sin(\omega_s t)$ with $\omega_s \sim \mathcal{U}(0.1, 0.5)$ Hz, combined with speed $\mathbf{u}_{\text{speed}}(t) = v_c + A\sin(\omega_v t)$, where $\omega_v \sim \mathcal{U}(0.1, 2.5)$ Hz. The speed amplitude $A$ and center velocity $v_c$ are determined such that the minimum speed lies in $\mathcal{U}(3, 4)$ m/s and maximum speed lies in $\mathcal{U}(8, 10)$ m/s. 

We adopt a gravity-aligned body frame, in which current state is denoted as
$
[0,\, 0,\, 0,\, \phi_t, \varphi_t,\, 0,\, \dot{\eta}_t,\, v_t]
\in \mathbb{R}^{8},
$
where the position components $(x, y, z)$ and yaw are zeroed (and therefore omitted), while $\phi_t$ and $\varphi_t$ retain their world-frame values. 
To better capture high-speed kinodynamics, we additionally include the yaw rate $\dot{\eta}_t$ 
and longitudinal velocity $v_t$.
The corresponding next state encodes the relative motion in the current body frame:
$
\mathbf{s}_{t+1} =
[\Delta x,\, \Delta y,\, \Delta z,\, \phi_{t+1},\, \varphi_{t+1},\, \Delta \eta]
\in \mathbb{R}^{6},
$
where $(\Delta x, \Delta y, \Delta z, \Delta \eta)$ denote body-frame pose change, 
and $(\phi_{t+1}, \varphi_{t+1})$ are the absolute roll and pitch angles at the next timestep.

\subsection{Cross-Vehicle Mobility Representation}
\label{sec::cv_mr}
The cross-vehicle mobility encoder consists of 4 Transformer blocks, each with embedding dimension $d=16$ and 4 attention heads. For each trajectory, transition tokens $\mathbf{x}_t = [\mathbf{s}_t, \mathbf{u}_t, \mathbf{s}_{t+1}] \in \mathbb{R}^{14}$ are linearly projected to $\mathbb{R}^{16}$ and prepended with a learnable CLS token augmented by sinusoidal positional embeddings.
Vehicle configuration parameters $(\alpha_m, \mu_f, \alpha_s)$ are first min-max normalized to $[0,1]$ using the predefined ranges: $\alpha_m \in [0.5, 4.0]$, $\mu_f \in [0.6, 0.9]$, and $\alpha_s \in [0.6, 1.8]$, and then embedded through a two-layer MLP with $\{3, 8, 16\}$ neurons and Tanh activations into a continuous $\mathbb{R}^{16}$ embedding space. The trajectory encoder and configuration modulator share the same Transformer backbone, with AdaLN conditioning applied exclusively at the final Transformer block with a scale factor of 0.5. 

To enable flexible inference, the configuration modulator adopts a learnable null embedding $\mathbf{e}_{\varnothing} \in \mathbb{R}^{16}$. During training, configuration embeddings are randomly replaced by $\mathbf{e}_\varnothing$ with a dropout probability of 0.1. This mechanism allows the encoder to operate in both \emph{conditional} mode, incorporating known physical configurations, and \emph{unconditional} mode, relying solely on trajectories when configuration information is unavailable.

The mobility encoder is trained with the dual-path triplet loss with margin $\delta=4.0$. We use the Adam optimizer with learning rate $1\times10^{-4}$ and batch size 128, training for up to 100K iterations. Early stopping with patience 10 is applied based on conditional embedding separation evaluated every 1,000 iterations on a held-out validation set.

\subsection{Kinodynamics Model}
The kinodynamics model $f_\theta$ is implemented as a lightweight MLP that predicts the 6-DoF state change from the current state and action. The model processes state and action inputs through separate branches, each implemented as a two-layer MLP with $\{8, 16\}$ hidden neurons, LayerNorm, and Tanh activations. The encoded features are concatenated and passed through two output heads: a translation and yaw head predicting $[\Delta x, \Delta y, \Delta z, \Delta \eta]$, and a roll-pitch head predicting $[\phi_{t+1},\, \varphi_{t+1}]$, each consisting of two fully connected layers with dimensions $\{32, 16\}$, BatchNorm, ReLU activations, and dropout. 

During rapid kinodynamics adaptation, $f_\theta$ is trained using weighted knowledge transfer with gradient regulation described in Section~\ref{sec::rapid_adapt}. The training loss is computed via autoregressive rollout over a horizon of $T_{\text{pred}}=16$ steps. Constraint gradients for gradient regulation are computed from $\mathcal{D}_{\text{new}}$ using three sampled trajectories, and refreshed every 100 gradient steps. The kinodynamics model
performs 2,000 gradient steps until convergence using the Adam optimizer with learning rate $1\times10^{-3}$, minimizing the MSE between predicted and ground-truth state changes.

\section{EXPERIMENTS}
We design our experiments to evaluate \our from three aspects: (1) cross-vehicle few-shot adaptation via identified mobility neighbors, (2) few-shot generalization compared with Model-Agnostic Meta-Learning (MAML)~\cite{finn2017model} and AnyCar~\cite{xiao2025anycar}, and (3) ablation studies of both the adaptation mechanisms and mobility encoder design.

\subsection{Cross-Vehicle Few-Shot Adaptation}
In this experiment, we evaluate the ability of \our to identify and leverage the most relevant mobility neighbors within the structured latent space for effective cross-vehicle knowledge transfer. We compare \our against two baselines:
\begin{itemize}
    \item \textbf{From Scratch:} A model trained on the full set of 400 trajectories collected from the new vehicle, serving as a data-intensive upper-bound; and
    \item \textbf{Mobility Neighbors:} Models trained from the identified mobility neighbors and directly applied to the new vehicle without any adaptation.
\end{itemize}


\begin{table}[h]
\centering
\caption{Cross-Vehicle Few-Shot Adaptation Performance on New Vehicle Configurations, denoted as $[\alpha_m, \mu_f, \alpha_s]$.}
\label{table:adapt}
\renewcommand{\arraystretch}{1.4}
\setlength{\tabcolsep}{6pt}
\small
\begin{tabular}{l l c c}
\toprule
\textbf{Configuration} & \textbf{Model} & \textbf{Dist.} & \textbf{MSE $\pm$ Std} $\downarrow$ \\
\midrule

\multirow{3}{*}{$[3.0,\,0.75,\,1.6]$}
& \cellcolor[gray]{.9}\our (Proposed) & \cellcolor[gray]{.9}-- & \cellcolor[gray]{.9}0.632 $\pm$ 0.825 \\
& Neighbor 1 & 0.0114 & 1.261 $\pm$ 1.117 \\
& \textbf{From Scratch} & -- & \textbf{0.173 $\pm$ 0.251} \\
\midrule

\multirow{4}{*}{$[0.6,\,0.7,\,0.8]$}
& \cellcolor[gray]{.9}\our (Proposed) & \cellcolor[gray]{.9}-- & \cellcolor[gray]{.9}0.734 $\pm$ 0.730 \\
& Neighbor 1 & 0.0752 & 2.241 $\pm$ 1.857 \\
& Neighbor 2 & 0.8167 & 3.305 $\pm$ 2.811 \\
& \textbf{From Scratch} & -- & \textbf{0.569 $\pm$ 0.775} \\
\midrule

\multirow{5}{*}{$[0.6,\,0.75,\,1.2]$}
& \cellcolor[gray]{.9}\our (Proposed) & \cellcolor[gray]{.9}-- & \cellcolor[gray]{.9}0.696 $\pm$ 0.972 \\
& Neighbor 1 & 0.4402 & 0.894 $\pm$ 0.722 \\
& Neighbor 2 & 0.5402 & 1.938 $\pm$ 3.110 \\
& Neighbor 3 & 0.6123 & 2.495 $\pm$ 2.680 \\
& \textbf{From Scratch} & -- & \textbf{0.222 $\pm$ 0.241} \\
\bottomrule
\end{tabular}
\end{table}

For the first new vehicle, only one mobility neighbor is identified within the cosine distance threshold in the mobility latent space, and \our applies regulated gradient updates solely on this neighbor's dataset. For the second and third configurations, two and three neighbors are identified respectively, with prediction error correlating monotonically with distance, nearer neighbors yield lower MSE. As shown in Table \ref{table:adapt}, \our outperforms direct transfer from individual mobility neighbors without adaptation across all new vehicle configurations, achieving up to a 67.2\% reduction in prediction error. Although the ``From Scratch'' upper-bound attains the lowest MSE by leveraging 400 platform-specific trajectories, \our only uses three trajectories. This result highlights the effectiveness of structured mobility representation in substantially reducing data requirements while maintaining strong adaptation performance.

\subsection{Cross-Vehicle Few-Shot Generalization}
We compare \our against MAML~\cite{finn2017model} and AnyCar~\cite{xiao2025anycar} to evaluate cross-vehicle generalization in the few-shot regime. Unlike MAML, which learns a platform-agnostic parameter initialization, and AnyCar, which trains a universal dynamics model across platforms, \our explicitly leverages physical vehicle configurations to identify and prioritize the most relevant mobility neighbors within the structured latent space. During adaptation, \our uses three trajectories from the new vehicle for mobility neighbor and weight identification, as well as gradient regulation, without fine-tuning on platform-specific data. 

\begin{table}[h]
\centering
\caption{Cross-Vehicle Few-Shot Generalization Compared with MAML and AnyCar.}
\label{table:scale}
\renewcommand{\arraystretch}{1.4}
\setlength{\tabcolsep}{8pt}
\small
\begin{tabular}{lcc}
\toprule
\textbf{Model} & \textbf{New Vehicle Data} & \textbf{MSE $\pm$ Std} $\downarrow$ \\ 
\midrule
\rowcolor[gray]{.9} \textbf{\our (Proposed)} & 3 Trajectories & \textbf{0.734 $\pm$ 0.730} \\
\midrule
\multirow{2}{*}{MAML} & 3 Trajectories & 3.507 $\pm$ 7.582 \\
 & 400 Trajectories & 1.329 $\pm$ 2.412 \\ 
\midrule
\multirow{2}{*}{AnyCar} & 3 Trajectories & 1.299 $\pm$ 1.535 \\
 & 400 Trajectories & 0.929 $\pm$ 1.089 \\ 
\bottomrule
\end{tabular}
\end{table}

Table~\ref{table:scale} shows \our outperforms both baselines under the same few-shot setting. With only three trajectories, \our reduces prediction error by 79.1\% and 43.5\% compared to MAML and AnyCar respectively. Remarkably, even when MAML and AnyCar are fine-tuned on 400 trajectories of the new vehicle, their prediction accuracy remains inferior to \our adapted with just three trajectories. These results demonstrate that structured mobility representations informed by physical vehicle configurations enable more data-efficient and robust cross-vehicle generalization than purely data-driven approaches.

\subsection{Ablation Studies}
We conduct ablation studies to evaluate the contribution of key components in \our's kinodynamics adaptation mechanism and mobility encoder design.

\subsubsection{Kinodynamics Adaptation Components}
Each variant removes one component from the full kinodynamics adaptation mechanism:
\begin{itemize}
    
    \item \textbf{\our}: The complete adaptation framework incorporating mobility neighbor identification (MN), weighted dataset aggregation (WD), weighted loss optimization (WL), and gradient regulation (GR);

    \item \textbf{\our w/o MN}: Removes mobility neighbor identification. Instead of selecting the most relevant mobility neighbors in the latent space, all vehicles in the training fleet are treated uniformly during adaptation;

    \item \textbf{\our w/o WD}: Removes weighted dataset aggregation. All identified mobility neighbor datasets are treated uniformly, eliminating selective emphasis on the most relevant mobility neighbors;

    \item \textbf{\our w/o WL}: Removes weighted loss optimization. Despite weighted dataset aggregation, their losses are treated equally during optimization without similarity-based weighting; and

    \item \textbf{\our w/o GR}: Removes gradient regulation via limited new vehicle data. The model is adapted solely using weighted datasets and loss without constraining gradient updates to preserve consistency with the new platform.

\end{itemize}

\begin{table}[h]
\centering
\caption{Ablation Study for Kinodynamics Adaptation Mechanisms.}
\label{table:abl_kino}
\renewcommand{\arraystretch}{1.4}
\setlength{\tabcolsep}{15pt}
\small
\begin{tabular}{lc}
\toprule
\textbf{Variant} & \textbf{MSE $\pm$ Std} $\downarrow$ \\ 
\midrule
\rowcolor[gray]{.9} \textbf{\our (Proposed)} & \textbf{0.734 $\pm$ 0.730} \\
\our w/o MN & 1.674 $\pm$ 2.791 \\
\our w/o WD & 1.729 $\pm$ 2.554 \\
\our w/o WL & 1.034 $\pm$ 1.367 \\
\our w/o GR & 1.173 $\pm$ 1.333 \\ 
\bottomrule
\end{tabular}%
\end{table}

As shown in Table~\ref{table:abl_kino}, each component of the adaptation mechanism contributes to \our's performance. Mobility neighbor identification and weighted dataset aggregation have the largest impact: their removal increases prediction error by 128\% and 135\%  respectively, indicating that effective cross-vehicle adaptation requires both identifying the most relevant mobility neighbors and weighting their contributions proportionally during knowledge transfer. Weighted loss optimization and gradient regulation provide complementary gains, with their removal degrading performance by 41\% and 60\% respectively, confirming that prioritizing kinodynamically similar sources and constraining parameter updates with limited new vehicle data are both essential for robust adaptation.

\subsubsection{Mobility Encoder Architecture}
We compare three encoder variants: our \emph{conditional} encoder, which applies AdaLN to modulate trajectory representations using physical configuration embeddings during training and inference; an \emph{unconditional} encoder, trained with AdaLN modulation but deployed when configuration information is unavailable during inference; and a \emph{simple} encoder that relies solely on trajectories during both training and inference.

\begin{table}[h]
\centering
\caption{Ablation Study for Encoder Design.}
\label{table:abl_encoder}
\renewcommand{\arraystretch}{1.4}
\setlength{\tabcolsep}{15pt}
\small
\begin{tabular}{lc}
\toprule
\textbf{Encoder Variant} & \textbf{MSE $\pm$ Std} $\downarrow$ \\
\midrule
\rowcolor[gray]{.9} \textbf{Conditional (Proposed)} & \textbf{0.734 $\pm$ 0.730} \\
\rowcolor[gray]{.9} Unconditional (Proposed) & 1.043 $\pm$ 1.242 \\
Simple & 1.797 $\pm$ 4.033 \\
\bottomrule
\end{tabular}%
\end{table}

Table \ref{table:abl_encoder} validates the architectural design of injecting physical priors via AdaLN conditioning. By explicitly grounding the mobility representation in vehicle configurations, the \emph{conditional} encoder achieves 29.63\% reduction in prediction error compared to the \emph{unconditional} counterpart. Furthermore, removing the conditioning architecture entirely from training (\emph{simple}) severely degrades performance.

\subsection{Physical Experiment} 
For real-world validation, we deploy \our on different open-source 1/10th-scale Verti-4-Wheeler platforms~\cite{datar2024toward}, with state information provided at 100\,Hz via a motion capture system. The training fleet comprises three distinct configurations: Rear Right Wheel Removed, Four-Wheeled, and Four-Tracked. As the new platform, we introduce a heavy-payload configuration with increased mass and altered inertia distribution, representing an embodiment shift relative to the training fleet. 

\begin{table}[h]
\centering
\caption{Physical Experiment Results.}
\label{table:phy_res}
\renewcommand{\arraystretch}{1.4} 
\setlength{\tabcolsep}{15pt}       
\small
\begin{tabular}{lc}
\toprule
\textbf{Model} & \textbf{MSE $\pm$ Std} $\downarrow$ \\ 
\midrule
\cellcolor[gray]{.9}\our (Proposed) & \cellcolor[gray]{.9}0.144 $\pm$ 0.186 \\
Neighbor 1 & 0.151 $\pm$ 0.184 \\
Neighbor 2 & 0.247 $\pm$ 0.188 \\
\textbf{From Scratch} & \textbf{0.111 $\pm$ 0.195} \\ 
\bottomrule
\end{tabular}%
\end{table}

To construct the kinodynamics knowledge base datasets, each training vehicle is driven using sinusoidal steering and velocity commands for 750 seconds per configuration. For rapid adaptation, \our transfers knowledge to the new heavy-payload platform. The physical experiment results in Table~\ref{table:phy_res} demonstrate that \our achieves the lowest prediction error compared with the identified mobility neighbors, reducing prediction error by 4.6\% relative to the best neighbor using only 45\,s of new trajectory data. Although training from scratch attains a lower error with 750\,s of platform-specific data, the small performance gap indicates that \our recovers most of the heavy-payload kinodynamics with minimal interaction data.

\begin{figure}[h]
    \centering
    \includegraphics[width=\columnwidth]{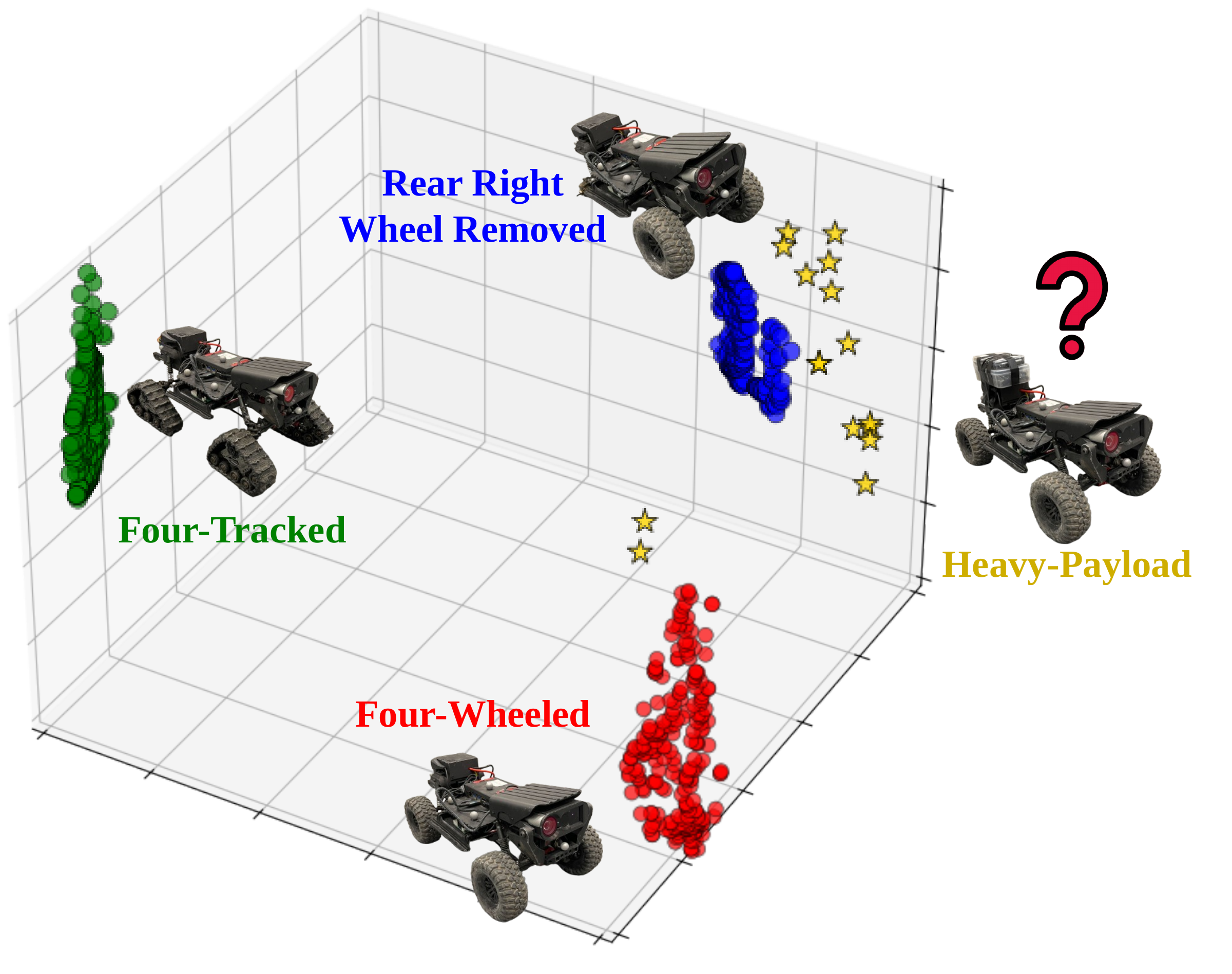}
    \caption{The physical mobility latent space is learned from three distinct configurations (blue, red, and green). When the new heavy-payload platform (yellow) is introduced, it is projected into this space to identify the most relevant mobility neighbors (blue and red) before rapid adaptation.} 
    \label{fig::phy_latent}
\end{figure}

The learned mobility latent space provides an interpretable explanation for the observed transfer behavior. As shown in Fig.~\ref{fig::phy_latent}, the heavy-payload configuration is embedded closer to the Rear Right Wheel Removed and Four-Wheeled configurations than the Four-Tracked platform, indicating an underlying kinodynamics similarity with these two mobility neighbors. This geometric structure is consistent with the performance trends shown in Table~\ref{table:phy_res}.

\section{Conclusions and Limitations}
We present \our, a novel framework designed to address the scalability challenges of enabling rapid kinodynamics adaptation across heterogeneous fleets. By leveraging a Transformer encoder conditioned with AdaLN, \our effectively maps vehicle trajectories and  physical configurations into a shared, structured mobility latent space. This representation enables reliable identification of the most relevant mobility neighbors for any newly introduced platform. Through weighted dataset aggregation, weighted loss optimization, and gradient regulation, the proposed framework achieves rapid kinodynamics adaptation using limited one minute of new data. Our evaluation demonstrates that \our reduces prediction error by up to 67.2\% compared to direct neighbor transfer, highlighting its effectiveness for scalable cross-vehicle mobility knowledge transfer.

A key limitation of this work is that it is evaluated only on flat terrain, where kinodynamics variations are primarily driven by physical configurations rather than diverse off-road terrain properties, obstacles, or dynamic environmental conditions. Future work will extend the mobility representation to jointly consider vehicle embodiment together with terrain geometry and semantics, enabling more comprehensive and realistic rapid kinodynamics adaptation.

\section*{Acknowledgments}
This work has taken place in the RobotiXX Laboratory at George Mason University. RobotiXX research is supported by National Science Foundation (NSF, 2350352), Army Research Office (ARO, W911NF2320004, W911NF2520011), Army Ground Vehicle Systems Center (GVSC), Google DeepMind (GDM), Microsoft Research (MSR), Clearpath Robotics, FrodoBots Lab, Raytheon Technologies (RTX), Tangenta, 4-VA, Mason Innovation Exchange (MIX), and Walmart.



\bibliographystyle{IEEEtran}
\bibliography{IEEEabrv,references}

\end{document}